\documentclass[11pt,a4paper]{article}
\usepackage{acl2019}
\usepackage{times}
\usepackage{latexsym}
\usepackage{multirow}
\usepackage[T1]{fontenc}
\usepackage{url}
\usepackage{subfigure}
\usepackage{graphicx}
\usepackage{amsmath}
\DeclareMathOperator{\cosine}{cosine}
\DeclareMathOperator{\avg}{avg}
\DeclareGraphicsExtensions{.eps,.ps,.jpg,.bmp,.png}

\aclfinalcopy 

\title{Do We Need Neural Models to Explain Human Judgments of Acceptability?}

\author{Wang Jing \\
  Beijing International Studies University \\
  Beijing, China \\
  {\tt wangjing1204@foxmail.com} \\\And
  M. A. Kelly \\
  Penn State \\
  University Park, PA, USA\\
  {\tt mak582@psu.edu} \\\And
  David Reitter \\
  Google Research \\
  New York, NY, USA \\
  {\tt reitter@google.com} \\}

\date{}

\begin{document}
\maketitle
\begin{abstract}
Native speakers can judge whether a sentence is an acceptable instance of their language. Acceptability provides a means of evaluating whether computational language models are processing language in a human-like manner. We test the ability of computational language models, simple language features, and word embeddings to predict native English speakers' judgments of acceptability on English-language essays written by non-native speakers. We find that much of the sentence acceptability variance can be captured by a combination of features including misspellings, word order, and word similarity (Pearson's $r=0.494$). While predictive neural models fit acceptability judgments well ($r=0.527$), we find that a 4-gram model with statistical smoothing is just as good ($r=0.528$). Thanks to incorporating a count of misspellings, our 4-gram model surpasses both the previous unsupervised state-of-the art (Lau et al, 2015, $r=0.472$), 
and the average non-expert native speaker ($r=0.46$). Our results demonstrate that acceptability is well captured by n-gram statistics and simple language features.
\end{abstract}

\section{Introduction}

Proficient language users, when given a sentence in their language, are able to judge the \emph{acceptability} of the sentence when asked whether the sentence is natural, well-formed, or grammatical. 
Acceptability is of interest to cognitive scientists because it provides a means for evaluating whether computational models of language are processing language in a manner similar to humans. 

At the same time, models of acceptability have applications in natural language processing. For example, they can be used to evaluate the fluency of machine translation outputs, of answers produced by question answering systems, and automatically generated language snippets quite generally \cite{lau2015unsupervised}. In computer-assisted learning, acceptability models can help grade essays and provide feedback to native speakers as well as language learners. Acceptability may also be a more precise training signal for general-purpose language models (albeit a costly-to-obtain one).

Recently, great progress has been made in language modeling. 
But do the models process language in a manner similar to humans? And are the models good at discriminating between acceptable and unacceptable language, or are the models merely sensitive to the distinction between probable and improbable language? Is acceptability statistical in nature, or best understood through the rules of a formal grammar?

In this paper, we investigate what models and language features can capture the acceptability judgements of native English speakers on sentences of English. We train different word embeddings, including \emph{Word2Vec} \cite[\emph{skip-gram} and \emph{continuous bag of words};][]{mikolov2013efficient,mikolov2013distributed}, the \emph{GloVe} model \cite{pennington2014glove}, and the \emph{Hierarchical Holographic Model} \cite[HHM;][]{kelly2017degrees}, to investigate the role of semantic features in acceptability. We train \textit{n}-gram models, simple recurrent neural network language models (RNN), and long-short term memory language models (LSTM), to obtain sentence-level probabilities. We explore how individual word frequency, \textit{n}-gram frequency, spelling errors, the order of words in sentences, and the semantic coherence of a sentence, contribute to human judgments of sentence acceptability.

Acceptability exhibits gradience \cite{keller2000gradience,lau2014measuring}. Accordingly, we treat acceptability as a continuous variable. We evaluate model performance by measuring the correlation between each model's acceptability prediction and a gold standard score based on human judgments. 

We are concerned with models that learn in an \emph{unsupervised} fashion. Supervised training requires copious labelled data and explicit examples of both acceptable and unacceptable sentences. Conversely, we use only the type of data that is most available to a human learner. Humans learn mainly through exposure to a language. While explicit training on what is and is not acceptable often occurs in school, this training is unnecessary for native language competence. 




In what follows, we discuss related work (\S\ref{Sec:Related}), provide an overview of our methodology (\S \ref{Sec:Method}), including models (\S\ref{Sec:MethodSemantic} \& \S\ref{Sec:MethodLM}), how we calculate an acceptability estimate from the models (\S\ref{Sec:Measure}), and data set (\S\ref{Sec:GUG}), before presenting the results of our experiments with word embeddings (\S\ref{Sentence Word Compatibility}) and language models (\S\ref{Sec:ExpLM}).

\section{Related Work}
\label{Sec:Related}
Prior work has focused on predicting sentence acceptability using syntactic parsing \cite{blache2006acceptability,Wong2010ParserFF,ferraro2012judging,clark2013statistical}, statistical language models \cite{clark2013statistical,lau2015unsupervised}, and linguistic features \cite{Heilman2014PredictingGO,clark2013towards}. Neural language models have shown evidence of acquiring deeper grammatical competence beyond mere \textit{n}-gram statistics \cite{Gulordava2018greenideas}, suggesting that the models are a good basis for modelling human acceptability judgements \cite{lau2015unsupervised}. 

 \citet{Vecchi2017spicy} uses word embeddings to predict human judgements of semantic acceptability on adjective - noun pairs (e.g., \textit{remarkable onion} vs. \textit{legislative onion}). Similarly, \citet{marelli2015affixation} compose morpheme embeddings to model the acceptability of novel word forms (e.g., \textit{undiligent} vs. \textit{unthird}). However, little work has been done on using distributional semantic models to predict sentence acceptability. 



\section{Methodology}
\label{Sec:Method}
There are many ways to determine to what extent a sentence is acceptable. Are the words spelled correctly? Are the words arranged in a correct order? Do the words create a coherent meaning?

\subsection{Semantic Coherence}
\label{Sec:MethodSemantic}
We hypothesize that if a sentence is acceptable to a native speaker, the words in the sentence will be compatible with each other with respect to meaning and to topic. Conversely, we hypothesize that if a word does not belong to a sentence, it is likely to be dissimilar in meaning to the other words. We call this measure \textit{semantic coherence}. Semantic coherence can be measured by computing the cosine similarity between the word and its context. The context can be the words near to the target word or all other words in the sentence.

\subsection{Language Models}
\label{Sec:MethodLM}
Language models predict probability distributions over sequences of words. Theoretically speaking, a language model can capture semantic coherence. \citet{lau2015unsupervised} demonstrate that a good language model can predict acceptability well. We improve upon \citeauthor{lau2015unsupervised}'s method for predicting acceptability  by also taking misspellings into account. 

\subsection{Acceptability Measures}
\label{Sec:Measure}
To estimate acceptability using language models or word embeddings, we take the log probability or cosine similarity produced by each and normalize it to the range [0,1] to get the model's \textit{Score}. We then convert the \textit{Score} into an acceptability measure using one of the methods in Table~\ref{measures}.

We introduce one more variable: misspellings (\emph{Mis}), which reflects the exponential effect of the count of misspellings. 


Because \citet{lau2015unsupervised} found that sentence acceptability is invariant with respect to sentence length or word frequency, we also normalize by sentence length and word frequency to produce a set of final scores based on the measures proposed by \citeauthor{lau2015unsupervised}. We normalize 
by word frequency in the \emph{NormMul} (normalized multiply) and \emph{NormSub} (normalized subtract) measures. The Syntactic Log Odds Ratio (SLOR) measure, proposed by \citet{Pauls2012LargeScaleSL}, accounts for both word frequency and sentence length. \citeauthor{lau2015unsupervised} found that \textit{SLOR} was the best predictor of acceptability. But, as we find that \textit{SLOR} does not perform well using our models, we
include 
\textit{NormMul}, and \textit{NormSub}, which are variants of the measures proposed by \citeauthor{lau2015unsupervised}. For each measure we compute its Pearson correlation coefficient with the sentence's gold standard to evaluate its effectiveness in predicting acceptability.

\begin{table}
    \renewcommand\arraystretch{2}
    \centering
    \begin{tabular}{ll}
         \bf Measures&\bf Equation  \\
    \hline
         Mis& $\mathit{Score}(\xi)^{(m+\alpha)}$\\
         NormMul&$\mathit{Mis}\times{P_u(\xi)}$\\
         NormSub&$\mathit{Mis}-P_u(\xi)$\\
         SLOR&$\frac{\mathit{NormSub}}{|\xi|}$\\
         
    \end{tabular}
    \caption{Measures for predicting the acceptability of a sentence. Notation: $Score$ is an estimate produced by a model (i.e., semantic coherence or sentence log probability), and normalized to [0,1]; $\xi$ is the sentence; ${|\xi|}$ is the sentence length; $m$ is the number of misspelled words in a sentence; $\alpha$ is a fitting parameter; $Pu(\xi)$ is the unigram probability of the sentence. We train the unigram model on English Wikipedia.}
    \label{measures}
\end{table}

\subsection{The GUG Dataset}
\label{Sec:GUG}
To test the ability of the models to predict acceptability, we need a collection of sentences that exhibit varying degrees of acceptability with ratings from native speakers. We use the Grammatical versus Ungrammatical (GUG) data set built by \citet{Heilman2014PredictingGO}. The GUG data set contains 3129 sentences randomly selected from essays written by non-native speakers of English as part of a test of English language proficiency (see Table \ref{Table:GUG}). \citet{Heilman2014PredictingGO} crowd-sourced acceptability ratings for the sentences on a 1 (incomprehensible) to 4 (perfect) scale, obtaining five ratings for each sentence. Each sentence also has an ``expert'' rating from a linguist. \citet{Heilman2014PredictingGO} randomly split the data into training (50\%), development (25\%), and test (25\%) sets. 

\begin{table*}[t!]
\begin{center}
\caption{Example sentences from \citet{Heilman2014PredictingGO}'s GUG data set with acceptability ratings.} 
\label{Table:GUG} 
\begin{tabular}{p{10cm}lll} 
\hline
sentence & expert & workers & mean \\
\hline
\textit{For not use car.} & 1 & [3, 4, 3, 4, 3] & 3.0 \\
\textit{I would like to initiate, myself , whatever I do on my trip to get much out of my trip.} & 2 & [4, 3, 2, 3, 3] & 2.8 \\
\textit{These kind of fish can't live so long in water that contain salt.} & 3 & [3, 3, 4, 3, 4] & 3.3 \\
\textit{So if you want me to choose right now, I will choose ordinary milk instead of that special kind.} & 4 & [3, 4, 3, 4, 4] & 3.7 \\
\hline
\end{tabular} 
\end{center} 
\end{table*}

In order to compare to the previous results in \citet{lau2015unsupervised}, we only use the GUG test set. We remove 23 sentences from GUG that have less than 5 words, lower-case all words, and extract 744 sentences for our test set. We take the average of the crowd-sourced ratings (across 5 workers) as the gold standard. To evaluate the models, we compute the correlation of the predicted ratings and the gold standard ratings. We correct misspelled words using the PyEnchant\footnote{\url{ http: //pythonhosted.org/pyenchant/}} spell-checker. We use PyEnchant's first suggestion as the corrected spelling. We count the number of misspelled words in every sentence, which serves as a feature for the \textit{Mis} measure (see Table \ref{measures}). 

To illustrate the difficulty of predicting acceptability, we compute the Pearson's correlation coefficient between the ratings of each human rater and the mean acceptability rating. We find that the correlations for crowd workers range from 0.440 to 0.485. 
However, the correlation between the expert and the average of all non-expert ratings is high, $r=0.753$. The high correlation could be an artifact of crowd worker selection. Crowd workers were screened using a qualifying test that assessed the agreement between their ratings and the expert ratings on a set of trial sentences, which could force a correlation. Conversely, the higher correlation for the expert and the average of the workers could reflect the expert's language expertise and the wisdom of the crowd. 

In sum, to achieve non-expert performance at predicting acceptability, computational models need a correlation to the mean acceptability rating of at least $r=0.440$, but to achieve expert performance may require a much higher correlation.

    
    

\section{Experiments}

\subsection{Acceptability as Semantic Coherence}
\label{Sentence Word Compatibility}
We hypothesize that more acceptable sentences have higher semantic coherence. We quantify semantic coherence between a word and its context as the cosine similarity between the vector representing the word and the vector representing the sentence without the word. The semantic coherence of the sentence as a whole is then computed as either the minimum or average of each word's similarity to the rest of the sentence.

To compute semantic coherence, we train word embeddings on two English language corpora: the British National Corpus \cite[\textbf{BNC}; 100 million words;][]{british2007british} and the Novels Corpus \cite[\textbf{NC}; 145 million words;][]{Johns2016b}. The word embeddings we train are Word2Vec (skip-gram, or SK, and continuous bag of words, or CBOW), GloVe, and the Hierarchical Holographic Model (HHM).

\subsubsection{Word2Vec}
Using CBOW, Word2Vec predicts the current word from a window of surrounding context words, while in SK, Word2Vec uses the current word to predict the surrounding window of context words \cite{mikolov2013efficient,mikolov2013distributed}. We use \emph{Gensim Word2vec} to train CBOW and SK models with $300$ dimensions and a window size of $5$.\footnote{Gensim Word2vec from: \url{https://radimrehurek.com/gensim/}}

\subsubsection{GloVe}
\emph{GloVe} takes aggregate word-word co-occurrence statistics from a corpus \cite{pennington2014glove} and performs a dimensional reduction on the co-occurrence matrix to produce a set of word embeddings. We build $300$ dimensional GloVe embeddings on the BNC and NC corpora\footnote{Our GloVe implementation comes from \url{https://github.com/stanfordnlp/GloVe}. Other parameter values of the GloVe vectors: $x_{max}=100; window\_size = 10; iter = 100$}.

\subsubsection{Hierarchical Holographic Model}

GloVe and Word2vec treat sentences as unordered sets of words. In English, word order conveys much of the meaning of the sentence, and is critical in constructing a grammatical sentence. To account for this, we include in our analysis the Hierarchichal Holographic Model \citep[HHM;][]{kelly2017degrees}, a model sensitive to the order of words in a sentence. HHM generates multiple levels of representations, such that higher levels are sensitive to more abstract relationships between words, improving the ability of the model to capture part-of-speech relationships \citep{kelly2017degrees}. 

We trained three levels of representations with 1024 dimensions and a context window of 5 words to the left and right of each target word. Number of dimensions, level, and window size are HHM's only parameters, and as such, HHM is comparable to the other word embeddings.

\subsubsection{Semantic Coherence}

The semantic coherence of a sentence is either:
\begin{gather}
    Score = \min(\cosine(\mathit{word}_i, \mathit{context}_i))
    \\
    Score = \avg(\cosine(\mathit{word}_i, \mathit{context}_i))
\end{gather}

\noindent where $\min$ is the minimum, $\avg$ is the average, $w_i$ is the $i$-th word's representation in a sentence 
and $context_i$ is the context representation of the sentence without $w_i$. If a word in the GUG test set is not in the corpus (169 test set words not in NC, 0 words not in BNC), we use a random embedding instead. We have two methods for computing the context representation $context_i$ of word $w_i$. One method is to sum:\\
\begin{equation}
    \mathit{context}_i = \frac{\sum{(w_1,...,w_{i-1},w_{i+1},...w_n)}}{n}
    \label{contextM}
\end{equation}
We can also get the context by building a holographic representation using HHM's environment vectors via a method described in \citet{Jones2007RepresentingWM} and \citet{kelly2017degrees}. By using aperiodic convolution to combine environment vectors into bigrams, trigrams, tetragrams, etc., up to \textit{n}-grams of sentence length, we can construct a representation of the sentence that accounts for the ordering of the words within it.

\subsubsection{Results}

Tables \ref{NOVEL compatibility results contextM} and \ref{BNC compatibility results contextM} show that semantic coherence, by itself, correlates weakly with acceptability when the sentence context is computed as a sum (as in Eqn. \ref{contextM}). Correlations for the \textit{Score} range from $r=-0.181$ (CBOW on BNC) to $r=0.185$ (SK on NC) with an average correlation across word embeddings and corpora of only $r=0.058$. 

We can improve the correlation by incorporating misspellings and unigram probability. GloVe is the best performing model when using the combined measures. We experimented with different values of the fitting parameter $\alpha$ for the \textit{Mis} measure and found the highest correlation for $\alpha=0$ and using the minimum (rather than average) semantic coherence of the sentence (GloVe, $r=0.449$ on NC and $r=0.446$ on BNC), which can be further improved using \textit{NormMul} by multiplying by the unigram probability (GloVe, $r=0.464$ on NC and $r=460$ on BNC). However, an $\alpha$ of zero indicates that the representations rely heavily on the misspellings to predict acceptability.

Conversely, when the context vector is computed holographically, such that the ordering of the words in the sentence is preserved, we find that the semantic coherence score is a stronger predictor of acceptability then when using a sum to construct the context (see Table \ref{compatibility results contextH}). \textit{Score} ranges from $r=0.140$ to $r=0.314$ and an average of $r=0.201$. By incorporating misspellings ($\alpha=0.3$), the correlation for average coherence increases to $r=0.471$ (NC) and $0.457$ (BNC) for Level 1 of HHM and to $0.467$ (NC) and $0.494$ (BNC) for Level 2 of HHM.

In sum, while semantic coherence is not a strong predictor of acceptability, when combined with sensitivity to the order of the words in the sentence and the number of misspellings, it can act as an effective predictor of acceptability ($r=0.494$).


\begin{table*}[t!]
    \centering
    \small
    \begin{tabular}{l|cccccccccccccc}
    \bf measure&
    \multicolumn{2}{c}{\bf HHM1}&
    \multicolumn{2}{c}{\bf HHM2}&
    \multicolumn{2}{c}{\bf HHM3}&
    \multicolumn{2}{c}{\bf CBOW}&
    \multicolumn{2}{c}{\bf SK}&
    \multicolumn{2}{c}{\bf GloVe}\\
    
         &min&avg&min&avg&min&avg&min&avg&min&avg&min&avg\\
    \hline
        Score&0.08&0.073&0.15&0.114&0.157&0.115&0.029&-0.176&0.185&-0.028&0.089&-0.071\\
        Mis&0.425&0.405&0.378&0.367&0.35&0.338&0.444&0.41&0.427&0.397&0.449&0.426\\
        NormMul&0.444&0.431&0.405&0.403&0.383&0.381&0.458&0.44&0.445&0.436&\bf 0.464&0.445\\
        NormSub&0.339&0.306&0.284&0.249&0.252&0.21&0.346&0.273&0.332&0.238&0.356&0.339\\
        SLOR&0.283&0.263&0.268&0.25&0.247&0.225&0.288&0.239&0.298&0.242&0.323&0.276\\
    \end{tabular}
    \caption{Pearson's $r$ between semantic coherence and acceptability with $\alpha=0$ for $\mathit{Mis}$. Semantic coherence computed using minimum and average similarity between word and context representations. Context representation obtained via Equation~\ref{contextM}. Word embeddings were trained on NC. Boldface indicates the best performing measure.}
    \label{NOVEL compatibility results contextM}
\end{table*}

\begin{table*}[t!]
    \centering
    \small
    \begin{tabular}{l|cccccccccccccc}
    \bf measure&
    \multicolumn{2}{c}{\bf HHM1}&
    \multicolumn{2}{c}{\bf HHM2}&
    \multicolumn{2}{c}{\bf HHM3}&
    \multicolumn{2}{c}{\bf CBOW}&
    \multicolumn{2}{c}{\bf SK}&
    \multicolumn{2}{c}{\bf GLOVE}\\
    
         &min&avg&min&avg&min&avg&min&avg&min&avg&min&avg\\
    \hline
        Score&0.076&0.071&0.155&0.111&0.155&0.113&0.035&-0.181&0.173&-0.029&0.064&-0.068\\
        Mis&0.422&0.404&0.380&0.367&0.351&0.337&0.445&0.41&0.425&0.399&0.446&0.426\\
        NormMul&0.442&0.430&0.407&0.402&0.385&0.381&0.459&0.440&0.444&0.437&\bf 0.460&0.445\\
        NormSub&0.334&0.306&0.285&0.249&0.252&0.21&0.348&0.275&0.327&0.243&0.351&0.339\\
        SLOR&0.282&0.262&0.269&0.249&0.249&0.225&0.29&0.24&0.301&0.248&0.318&0.276\\
    \end{tabular}
    \caption{As in Table~\ref{NOVEL compatibility results contextM}, but with word embeddings trained on BNC.}
    \label{BNC compatibility results contextM}
\end{table*}


\begin{table*}[t!]
    \centering
    \small
    \begin{tabular}{cc}
    \begin{tabular}{l|llllll|}
         \bf measure&
    \multicolumn{2}{c}{\bf HHM1\_NC}&
    \multicolumn{2}{c}{\bf HHM2\_NC}&
    \multicolumn{2}{c|}{\bf HHM3\_NC}\\
    
         &min&avg&min&avg&min&avg\\
    \hline
        Score&0.153&0.198&0.19&0.27&0.191&0.232\\
        Mis&0.454&0.471&0.459&0.467&0.447&0.462\\
        NormMul&0.457&\bf 0.476&0.453&0.462&0.436&0.455\\
        NormSub&0.346&0.36&0.365&0.38&0.363&0.378\\
        SLOR&0.239&0.176&0.243&0.204&0.293&0.219\\ 
    \end{tabular}
    \begin{tabular}{llllll}
         
    \multicolumn{2}{c}{\bf HHM1\_BNC}&
    \multicolumn{2}{c}{\bf HHM2\_BNC}&
    \multicolumn{2}{c}{\bf HHM3\_BNC}\\
    
         min&avg&min&avg&min&avg\\
    \hline
        0.097&0.129&0.217&0.314&0.14&0.278\\
        0.435&0.457&0.475&\bf 0.494&0.425&0.482\\
        0.438&0.459&0.463&0.479&0.432&0.464\\
        0.308&0.318&0.381&0.404&0.274&0.409\\
        0.131&0.096&0.183&0.169&0.155&0.24\\
    \end{tabular}
    \end{tabular}
    \caption{Context representations obtained via the holographic method for HHMs. Measures computed with $\alpha=0.3$. Boldface indicates the best-performing measure.}
    \label{compatibility results contextH}
\end{table*}

\subsection{Acceptability using Language Models}
\label{Sec:ExpLM}
Language models can predict the probability of a sequence of words. \citet{lau2015unsupervised} compared acceptability judgments against predictions by several language models including \textit{n}-gram models, variations of Hidden Markov Models, latent Dirichlet allocation, a Bayesian chunker, and a recurrent neural network language model (RNNLM).
\citet{lau2015unsupervised} obtained acceptability scores using crowd-sourcing and the corpus of sentences was obtained using round-robin machine translation from English to a second language, and back.  On this (unpublished) data, the neural language model (RNNLM) performed best.

However, a 4-gram model trained on the British National Corpus beat the RNNLM when tested on the GUG test data \cite[$r=0.472$;][]{lau2015unsupervised}. 

We take the 4-gram model's performance (\(r=0.472\)) as the baseline in the following experiments. The primary difference between the translation dataset and the GUG dataset is that sentences in the translation dataset are produced by Google Translate automatically, while sentences in the GUG dataset are produced by a non-native speaker. We suspect that a language model may perform better on a dataset produced by machine translation than on a naturalistic dataset comprised of essays from L2 speakers, which may contain greater linguistic variability.

\subsubsection{Training}
\label{LM training}

We produce lexical 4- and 5-gram language models using Kneser-Ney smoothing \cite{stolcke2002srilm}, and a basic RNNLM 
\footnote{We use the \citet{mikolov2010rnnlm} implementation for the RNNLM: \url{ http://www.fit.vutbr.cz/~imikolov/rnnlm/}. Meta-parameter values of the RNNLM included the following: number of classes = 550; bptt = 4; bptt-block = 100, hidden = 600.} 
\cite{elman1998generalization, mikolov2012statistical}, on two large corpora of ``acceptable'' English:
the British National Corpus \citep[\textbf{BNC}: 130K unique words, 100M tokens;][]{british2007british} and on a corpus of novels \citep[\textbf{NC}: 39K unique words, 145M tokens;][]{Johns2016b}. We also train the RNNLM and an LSTMLM on NC using Tensorflow. The embedding layer of the models is initialized either to a Gaussian distribution of values with 1024 dimensions or a pre-trained embedding with 3072 dimensions (we use a concatenation of HHM1, HHM2 and HHM3 here, details about HHM are in section~\ref{Sentence Word Compatibility}). We set the LSTM's hidden layer size to 1024, the projection layer size to 128, and the maximum sequence length to 25.

The corpora are tokenized using NLTK\footnote{https://www.nltk.org} and all words are converted to lower case. In the GUG test set, we replace out-of-vocabulary words (i.e., words not in the corpora) and low-frequency ($<5$) words with the \verb|<unk>| signature. Sentences with less than $8$ words are also removed.


\subsubsection{Results}
\label{Sec:LMresult}

Score\_C in Table \ref{ngram correlation} equals the log probability of sentences in the spell-corrected test data and Score\_O is the log probability of sentences in the original test data.

Setting the \textit{Mis} score's fitting parameter $\alpha=1.3$ (see Table \ref{measures}) yielded best results for the lexical 4-gram model on BNC. The 4-gram model's correlation of $r=0.528$ improves notably upon Lau et al.'s best correlation of $0.472$. The RNNLM trained on NC performs similarly, at $r=0.527$.



Table~\ref{tab:LSTM correlation} shows the performance of the LSTM and RNN trained on NC with an embedding layer. We find that the best performing model is the LSTM with a pre-trained embedding layer that is not fixed and can be trained further. 

\emph{Mis} works well for capturing misspelled words. Table~\ref{ngram correlation} shows that the correlation of \emph{Mis} exceeds the original log probability no matter whether the log probability is computed on corrected test data or the original test data. The \textit{Mis} score's high correlation demonstrates that accounting for misspellings allows for a more accurate translation of log probability to acceptability. 

Unlike \citet{lau2015unsupervised}, who found \textit{SLOR} to be the best measure, we find that \emph{NormSub} performs better than other measures across all models (Table~\ref{ngram correlation}). \emph{NormSub} removes the influence of unigram probability from the acceptability score. Word frequency has little influence on acceptability ($r=0.2$ between acceptability and unigram probability), but greatly affects the log probability computed by the language models.

\begin{table*}[t!]
    \centering
    \small
    \begin{tabular}{l|cccccc}
    \multirow{2}{*}{\bf measure}&
    \multicolumn{2}{c}{\bf 4-gram}&
    \multicolumn{2}{c}{\bf 5-gram}&
    \multicolumn{2}{c}{\bf RNN}\\
    &BNC&NC&BNC&NC&BNC&NC\\
    \hline
    Score\_O&0.284&0.250&0.252&0.311&0.337&0.284\\
    Score\_C&0.315&0.302&0.315&0.302&0.295&0.314\\
    Mis&0.465&0.460&0.464&0.459&0.448&0.461\\
    NormMul&0.424&0.418&0.423&0.417&0.410&0.421\\
    NormSub&\bf 0.528&0.518&0.527&0.518&0.510&\bf 0.527\\
    SLOR&0.469&0.457&0.474&0.460&0.472&0.485\\
    \end{tabular}
    \caption{Pearson's $r$ between model predictions and mean acceptabiliy for language models trained on the BNC and NC corpora with $\alpha=1.3$. Boldface indicates the best performing models and measures.}
    \label{ngram correlation}
\end{table*}

\begin{table*}[h!]
    \centering
    \small
    \begin{tabular}{c|cccc}
    {\bf Measures}&{\bf LSTM} &{\bf LSTM}&{\bf RNN}  &{\bf LSTM}  \\
     &HHM (fixed)&HHM (trainable)&Gaussian&Gaussian\\
    \hline
Score&0.42&0.408&0.307&0.344\\
Mis&0.493&0.49&0.418&0.445\\
NormMul&0.453&0.447&0.385&0.406\\
NormSub&0.509&{\bf 0.522}&0.453&0.489\\
SLOR&0.38&0.4&0.339&0.374\\
    \end{tabular}
    \caption{Pearson's $r$ between model predictions and mean acceptability for language models trained on the NC corpus with $\alpha=2.1$. The models are an LSTM with fixed embeddings, an LSTM with trainable embeddings, an RNN with randomly initialized trainable embeddings, and an LSTM with randomly initialized trainable embeddings. Boldface indicates the best performing models and measures.}
    \label{tab:LSTM correlation}
\end{table*}

To test the robustness of our methodology on the test data, we also test the models on the GUG's development set (747 sentences after preprocessing) and find that a 5-gram model trained on the BNC gets the best correlation at \(r=0.467\).

\subsubsection{Perplexity versus Acceptability}
The perplexity of a language model on test data is commonly used to evaluate model performance. In this section we explore the relationship between perplexity and acceptability.

We trained several neural network language models using Tensorflow including RNNLM, GRULM and LSTMLM on the Penn Tree Bank corpus (PTBC)\footnote{\url{https://catalog.ldc.upenn.edu/LDC99T42}}. PTBC is smaller than BNC and NC with a 10k vocabulary size, so it is faster to train on. We set the word embedding size to 300, initializing to random values sampled from a Gaussian Distribution, and set the embeddings to be to trainable in the model. We set the maximum sequence length to 25 and hidden, recurrent layer size to 1024. Model training halts if validation loss does not decrease for three consecutive epochs and the perplexity (ppl) on the PTBC test data is used to evaluate models in the last epoch.


\begin{figure}[h!]
    \centering
    \includegraphics[width=7cm]{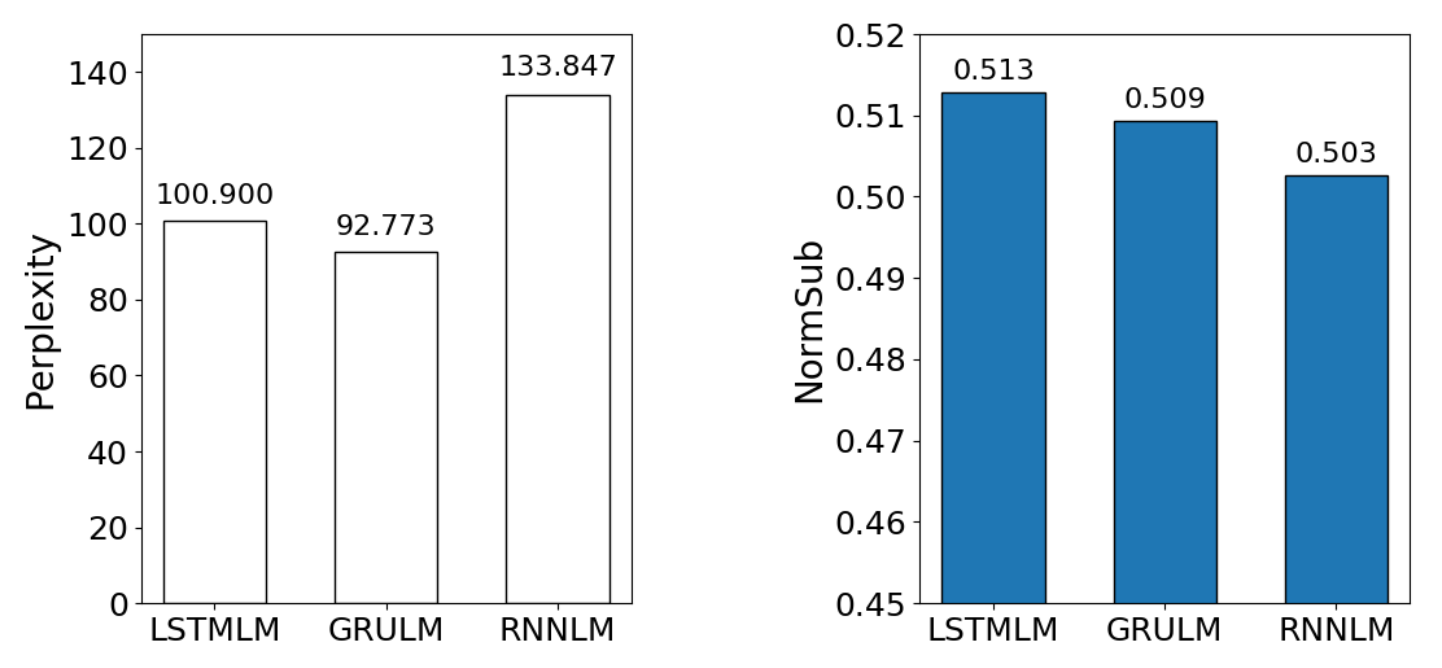}
    \caption{Perplexity versus correlation of \textit{NormSub} to acceptability across different models.}
    \label{fig:grammaticality vs test ppl}
\end{figure}

Over the course of training, as the perplexity on the validation data decreases, we find that the correlation to acceptability tends to increase. However, when comparing perplexities across different types of models, the perplexity is not strongly predictive of which type of model makes more human-like acceptability predictions, as shown in Figure \ref{fig:grammaticality vs test ppl}. Even though the GRULM has the lowest perplexity of the three models (92.773), the LSTMLM has the highest correlation to human acceptability ($r=0.513$). 
Thus, a different model with a different architecture and a lower perplexity may not be better able to predict acceptability.

\section{Discussion}

Modeling acceptability provides a window into the human brain's language engine. We explore what aspects of language are important in order to account for what people consider acceptable, well-formed sentences. In particular, we examine the importance of misspellings, semantic coherence, and \textit{n}-gram probability. To examine each, we use simple features, word embeddings, \textit{n}-gram models, and predictive neural language models.

We find that accounting for misspellings considerably improves the ability of unsupervised models to capture acceptability judgments. Prior work by \citet{Heilman2014PredictingGO} with supervised models found that the number of misspelled words was an important feature for their models. We develop a technique for incorporating misspellings into unsupervised models by correcting all spelling errors so that the model works with clean data, and then raising the model's acceptability estimate to the power of the number of misspelled words.

Using word embeddings to compute the semantic coherence of a sentence, we examine the contribution of semantic coherence with respect to acceptability. By itself, semantic coherence correlates poorly with acceptability, providing evidence that meaning constitutes only a small part of what makes a sentence acceptable or unacceptable. The finding that semantics plays only a small role in acceptability is in keeping with \citet{Chomsky1956}'s arguments for a distinction between syntactically well-formed sentences and semantically well-formed sentences.

However, when semantic coherence is combined with misspellings and unigram probabilities (i.e., word frequencies), we can account for much of the variability in acceptability ($r=0.46$). We can further improve the correlation by incorporating the order of the words in the sentence into our measure of semantic coherence ($r=0.49$). These results suggest that acceptability is not wholly independent of semantics.

Language models provide a means of estimating the probability of a sentence, which in turn can be used to predict acceptability. We replicate prior work \cite{lau2015unsupervised} in finding that the RNNLM is a good language model for accounting for acceptability ($r=0.53$). Yet, a simple 4- or 5-gram model with statistical smoothing is just as good as an RNNLM.  So, 4- or 5-gram frequencies may be what the neural network model primarily learns to rely upon to model human language with respect to acceptability judgments.

The quality of a language model is typically evaluated using perplexity, a measure of the uncertainty of the model's predictions. We find that, unsurprisingly, over the course of training, as the language model's perplexity decreases, the model's ability to predict acceptability increases. However, the decreases in perplexity across language model architectures do not directly map onto increases in correlation to human acceptability judgements. 

We replicate \citet{lau2015unsupervised}'s finding that the log probabilities produced by language models need to be normalized by subtracting the unigram probability. This normalization, \textit{NormSub}, prevents the models from underestimating the acceptability of sentences with low frequency words. \citet{lau2015unsupervised} found that the Syntactic Log Odds Ratio \cite[SLOR;][]{Pauls2012LargeScaleSL}, which further normalizes by dividing by sentence length, proved the best method for converting the log probabilities of language models into an acceptability prediction. However, we consistently found that \textit{SLOR} performed less well at predicting acceptability than \textit{NormSub} (see Tables \ref{ngram correlation} and \ref{tab:LSTM correlation}).

The difference between our findings and \citeauthor{lau2015unsupervised}'s may be attributable to the very different datasets we use: \citeauthor{lau2015unsupervised} primarily report results for a machine translation dataset, whereas we evaluate results on a dataset of sentences produced by second language speakers. The appropriateness of normalizing by sentence length thus may be dependent on the characteristics of the language being evaluated for acceptability.

In summary, the best and simplest model of acceptability judgment needs only three things: 4- or 5-gram frequency (with statistical smoothing), misspellings, and word (or unigram) frequency. 

Accounting for acceptability does not require a complex model of syntax or even a sophisticated predictive neural model.  Rather, acceptability appears to be a function of correct spelling, the statistical relationships between words, word meanings, and word order. Strictly `grammatical' language is not necessarily idiomatic, and NLP methods 
may want to strive for idiomatic and acceptable rather than merely grammatical language.

The correlation of non-expert of human raters to mean acceptability ranges from 0.440 to 0.485. Our models' correlation with the mean acceptability exceeds the rater reliability.

Yet even our best model is not as good as the expert rater, which has a correlation of $0.75$ with the mean rating. This is an obvious challenge for future work \citep[cf.,][]{dkabrowska2010naive}. Do experts have more experience and thus, different assumptions of lexical or syntactic distributions, or do they just interpret the task differently, perhaps isolating grammaticality from the less well-defined acceptability? Do their judgments represent less dialectal variety, or are they correlated with ease-of-processing?  Future models will, hopefully, be able to capture this expert-level performance. 

We have considered only unsupervised models of acceptability on the assumption that unsupervised data best reflects the learning environment of the average native speaker of the language. However, supervised approaches may be appropriate for modelling expert performance, as language experts have likely had the benefit of explicit training on what is and is not acceptable and `proper' language. Though it remains an open question whether capturing such expertise is even desirable given that `proper' language is typically associated with a specific region or class, rather than the population as a whole \cite[][pp. 984-985]{McArthur1992}.

\section{Conclusion}


While more sophisticated neural language models may have lower perplexity than simpler language models, we find that simple \textit{n}-gram models perform just as well at predicting acceptability. By incorporating a count of misspellings, our 4-gram model surpasses the previous unsupervised state-of-the art (Lau et al, 2015, $r=0.472$), reducing the gap to expert performance ($r=0.75$) and surpassing the average non-expert native speaker ($r=0.46$). It is also possible to achieve a high correlation to acceptability without using a predictive language model at all, by combining number of misspellings, semantic similarity between words, and word order information ($r=0.494$).

Our results suggest that the human ability to judge whether or not a given sentence is acceptable may be derived from sensitivity to simple, statistical language features, without necessarily requiring syntactic rules or even a particularly complex machine learning model.

\section*{Acknowledgments}
This work was supported by a Post-Doctoral Fellowship from the Natural Sciences and Engineering Research Council of Canada (NSERC) to M. A. Kelly and a National Science Foundation grant (BCS-1734304) to David Reitter and M. A. Kelly.

\bibliography{emnlp2018}
\bibliographystyle{acl_natbib}

\end{document}